\documentclass{svproc}
\usepackage{url}
\usepackage[ruled, lined, linesnumbered, longend]{algorithm2e}
\usepackage{graphicx}
\usepackage{amsmath}
\usepackage{amssymb}
\usepackage{multirow}
\usepackage{xcolor}

\begin{document}
\mainmatter              
\title{Rethinking Data Input for Point Cloud Upsampling}
\titlerunning{Hamiltonian Mechanics}  
%
\author{Tongxu Zhang\inst{1}}
\authorrunning{Tongxu Zhang} 
%
\tocauthor{Tongxu Zhang}
\institute{The Hong Kong Polytechnic University, Hung Hom, Hong Kong, China\\
\email{tozhang@polyu.edu.hk}}

\maketitle              

\begin{abstract}
Point cloud upsampling is crucial for tasks like 3D reconstruction. While existing methods rely on patch-based inputs, and there is no research discussing the differences and principles between point cloud model full input and patch based input. Ergo, we propose a novel approach using whole model inputs i.e. Average Segment input. Our experiments on PU1K and ABC datasets reveal that patch-based inputs consistently outperform whole model inputs. To understand this, we will delve into factors in feature extraction, and network architecture that influence upsampling results. \dots
\end{abstract}
\section{Introduction}
The improvement of 3D scanners in 3D sensors, as well as the rapid development of deep learning in computer vision, have led to a gradual shift in the popularity of computer vision technology from 2D image to 3D model. Especially in the direction of autonomous driving and robots that exchange with the physical world, there is a particular focus on three-dimensional vision \cite{endres20133,roriz2021automotive,naseer2018indoor,zhou2018voxelnet}.
\begin{figure}[tb]
  \centering
  \includegraphics[height=3.6cm]{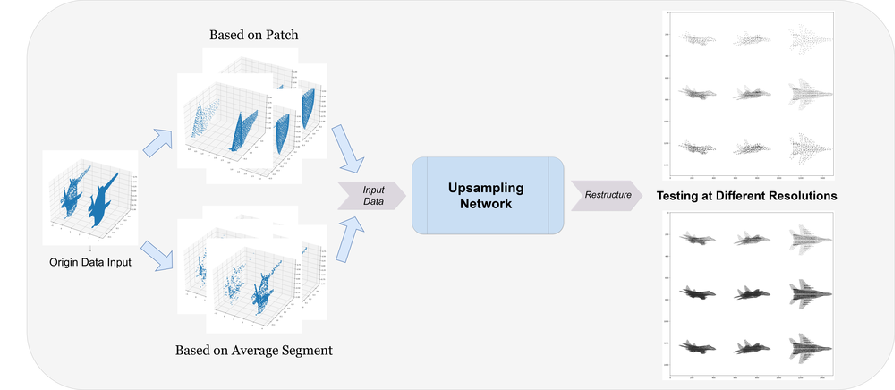}
  \caption{The original Mesh models have been transformed into a point cloud through patch based sampling and the Average Segment method proposed in this article.}
  \label{fig:1}
\end{figure}
Vis-a-vis other representations such as Mesh, point clouds are an effective way to represent three-dimensional shapes that are widely used in the aforementioned fields. In addition, the task of reverse engineering CAD models from 3D point clouds is receiving increasing attention \cite{alai2013review}. In the process of redesigning the model obtained from reverse engineering, although there are Application Programming Interface (API) available in CAD software \cite{kao1998development}, if these designs involve reverse engineering and highly customized and personalized designs, CAD APIs are powerless \cite{shatnawi2017reverse}. Therefore, the generation of model point clouds is crucial.

The point cloud upsampling technology based on deep learning has developed with the development of deep learning, extracting multi-resolution features from the initial PU-Net \cite{yu2018punet} on the basis of pointnet++ \cite{qi2017pointnet}. Then, multiple independent multi-layer perceptrons (MLPs) are introduced to generate the required dense point cloud. Moving on to MPU/3PU \cite{yifan2019patchbased}, KNN is used for interpolation and EdgeConv \cite{wang2019dynamic} is used to extract features for hierarchical reconstruction through recursive upsampling. Implementing upsampling PU-GAN \cite{li2019pugan} based on GAN \cite{goodfellow2020generative} generation architecture, guiding the generator to upsample and approximate the real point cloud. PU-GCN \cite{qian2021pugcn} modifies the upsampling module from simple duplicate to Nodeshuffle based on feature extraction in MPU, achieving very good results. And Transformer's point cloud upsampling network \cite{qiu2022putransformer,vaswani2017attention}. There are also methods to create coarse point clouds and then refine them by combining global and local methods \cite{li2021point}.

The above methods mainly utilize Euclidean spatial features at potential spatial points, upsampling from low resolution to high resolution. The inherent limitations have also been addressed by the work of upsampling point clouds at any scale \cite{zhao2022selfsupervised,he2023gradpu} and capturing local information \cite{kim2024puedgeformer}.

Nevertheless, these works still have limitations, which are on the training dataset. Existing point cloud upsampling work \cite{qian2021pugcn,qiu2022putransformer,zhao2022selfsupervised,he2023gradpu,kim2024puedgeformer,mao2022pu} without exception used the PU1K dataset \cite{qian2021pugcn}, which was used based on the methods of MPU and PU-GAN \cite{yifan2019patchbased,li2019pugan}. To obtain training data by randomly sampling the entire 3D model using Patch under conditions of limited computing power, especially limited graphics memory. Undoubtedly, Patch based data input can reduce memory consumption and computational complexity, but it can bring about boundary effects and continuity loss in reverse engineering. It is necessary to explore a new point cloud input method to avoid the above issues. Moreover, the existing point cloud upsampling work focuses on 3D models similar to the PU1K dataset, while neglecting the exploration of point cloud upsampling for parts and CAD. In addition to conducting experiments on the PU1K dataset, this article will also discuss the upsampling effect of point cloud models on workpieces on the ABC dataset \cite{koch2019abc}.

Based on the above issues, in order to explore why numerous point cloud sampling data are still based on Patch input, this paper proposes a new data input method. Unlike dividing point clouds according to Patch, this article takes the point cloud model uniformly and randomly divided and corresponds them one by one as input, which is named as Average Segment (AS). After verifying the method proposed in this article, experiments were conducted on point cloud upsampling using the PU-GCN \cite{qian2021pugcn} network, and cross validation was performed on both the PU1k and ABC datasets. And during the experiment, the encoder and decoder of PU-GCN \cite{qian2021pugcn} were ablated.

To solve the above problems, this paper proposes a new data input method. Unlike dividing point clouds according to patches, this paper uniformly and randomly divides point cloud models and uses them as inputs one by one. After verifying the method proposed in this article, experiments were conducted on point cloud upsampling using the PU-GCN \cite{qian2021pugcn}, and cross validation was performed on both the PU1k \cite{qian2021pugcn} and ABC datasets \cite{koch2019abc}. And during the experiment, the encoder and decoder of PU-GCN \cite{qian2021pugcn} were ablated.

The contributions of this article are as follows:

-Proposing a new data input method for point cloud upsampling, namely the Average Segment (AS) method. In theory, the boundary effects and continuity issues brought about by two patch and uniform segmentation methods were discussed.

-Verifying the effectiveness of Patch based method through experiments with AS method and explains the reasons why patch input is superior to AS input.

-Conducting ablation experiments on the main structure of PU-GCN, determining that upsampling mainly relies on the extended feature modules in the decoder, and pointing out the key link in point cloud generation.

\section{Methodology}

\subsection{Overview}
As shown in Figure ~\ref{fig:1}, given a sparse point cloud $\mathcal{P}\in\mathbb{R}^{N\times3}$, the input methods based on Patch and the AS method proposed in this paper are used as data inputs, with PU-GCN as the network backbone, to generate a dense point cloud $\mathcal{P'}\in\mathbb{R}^{rN\times3}$, where $r$ is the upsampling scale. The reason for choosing PU-GCN as the backbone of the network is that MPU and PU-GCN have similar feature extraction methods. Both use KNN to construct a graph of the input point cloud, with nodes being points and edges defining $k$ nearest point neighborhoods. Then, EdgeConv uses these distances to calculate the edge features of each point. The subsequent PU-Transformer in the upsampling module adopted the Nodeshuffle proposed by PU-GCN, which forms a dense feature map compared to MPU's Duplicate.

Incidentally, where the input training data of MPU and PU-GAN, and the PU1k dataset proposed by PU-GCN both convert an entire mesh into a point cloud for upsampling and divide the mesh into multiple patches, and then upsample each patch separately. Accordingly, this entire point cloud has its own advantages and disadvantages as two methods for training and sampling into patches.

{\bf The disadvantage of converting the entire mesh into a point cloud for upsampling:} It may consume a lot of memory resources, especially for complex high-resolution meshes. This may result in insufficient memory or slow processing speed. And for computational complexity, converting large meshes directly into point clouds requires a significant amount of computing resources and time. When performing point cloud upsampling, this high computational complexity will further increase. In addition, the point cloud converted from the entire mesh may be very dense, which may contain a large number of unnecessary redundant points. This brings about the issue of density, as these redundant points increase computational complexity and may lead to overfitting of the model.

Ergo, to avoid memory consumption and computational complexity, the previously proposed patch based end-to-end training has emerged. Dividing the mesh into multiple patches for upsampling is more practical and efficient. But there are also disadvantages to this.

{\bf The disadvantage of dividing the mesh into multiple patches for upsampling:} Frankly, information is missing. Dividing the mesh into multiple patches may result in the loss of important local features, as each patch may not contain complete information about the entire object. This may affect the quality of upsampling. In addition, due to the existence of boundaries between patches, this brings about boundary effects. When dividing the mesh into multiple patches for upsampling, discontinuity or defects may occur. And therefore, the results sampled on each patch may not be continuous enough, as they are processed independently. This may affect the continuity and consistency of the final point cloud.

The AS i.e. {\bf A}verage {\bf S}egment method proposed in this article converts the entire mesh into a point cloud and performs uniform random sampling and segmentation, which can partially avoid some shortcomings of dividing the mesh into multiple patches. This is because we retained the overall local features in the point cloud through random sampling, and avoided boundary effects and information loss caused by dividing the mesh into discrete segments.

{\bf Whereas actually, there are still some potential disadvantages:} especially in terms of continuity. Although the overall local features are retained during sampling, point cloud data may not be continuous enough in some areas due to random sampling, and further processing may be needed to ensure the continuity and consistency of the point cloud.


\subsection{Data Input Based on Patch}
In this case, it is necessary to divide the Mesh into multiple patches. Assuming the entire Mesh contains $M$ vertices, divide it into $K$ patches. $\mathcal{P}_k$ contains $M_k$ vertices, where $\sum_{k=1}^{K} M_k \geq M$. Convert each patch to a point cloud and perform upsampling. After converting the vertices of each patch to a point cloud, it is assumed that the point cloud contains $N_k$ points and we want to upsample them into $N_k'$ points, typically $N_k'>N_k$.

Process each patch using a graph convolutional network, and record the output of the graph convolutional network as $\mathbf{h}_i$represents the feature representation of the $i$-th point. Then, use NodeShuffle technique to generate more points.

Let the graph convolutional network have $L$ layers, and the weight matrix of each layer is $\mathbf{W}_l$, then the output of layer $l$is:

\[
\mathbf{H}^{(l)} = \sigma(\mathbf{A} \mathbf{H}^{(l-1)} \mathbf{W}^{(l)})
\]

where:

$--$ $\mathbf{A}$ is an adjacency matrix.

$--$ $\mathbf{H}^{(0)}=\mathbf{X}$ representing the feature matrix of the input point.

$--$ $\sigma$ is the activation function.

And, we obtain a new feature representation $\mathbf{H}^{(L)}$and use NodeShuffle to generate a new set of points:

\[
\mathcal{P}_k' = \text{NodeShuffle}(\mathcal{P}_k)
\]

Therefore, there is an upsampled point cloud set, which concatenates all patch upsampled point cloud sets into a complete point cloud:

\[
\mathcal{P}' = \bigcup_{k=1}^{K} \mathcal{P}_k'
\]

\subsection{Data Input Based on Average Segment}
We can assume that the distribution of sampling points follows a uniform distribution. Assuming that the range of a point cloud is a bounded three-dimensional spatial area, we can use uniform distribution to simulate the sampling process of points.

After converting the entire mesh to a point cloud, uniform random sampling is performed. Assuming we want to randomly sample $M$ points from $N$ points as a subset, where $M$ is much smaller than $N$. The probability of each point being selected is $p = \frac{M}{N}$.

Assuming we have conducted $N$ independent Bernoulli experiments with a probability of success of $p$, then the number of successful experiments $X$ follows a binomial distribution $B (N, p)$. The selected number of points $S$ follows a binomial distribution $B (N, p)$:

\[
S = \sum_{i=1}^{N} X_i, \quad S \sim B(N, p)
\]

Selected point set $\mathcal{P}_{\text{sampled}}$ can be expressed as:

\[
\mathcal{P}_{\text{sampled}} = \{ \mathbf{p}_i \mid X_i = 1 \}
\]

Once sampling is completed, obtain a point set containing $M$ points $\mathcal{P}_{\text{sampled}}$. We hope to divide it into $K$ subsets, each containing $\frac{M}{K}$ points.

The $k$-th subset is denoted as $\mathcal{P}_k$, where $k = 1, 2, \ldots, K$. Assuming that $M$ can be divided by $K$, the range of points for each subset is:

\[
\mathcal{P}_k = \{ \mathbf{p}_i \mid (k-1) \cdot \frac{M}{K} + 1 \leq i \leq k \cdot \frac{M}{K} \}
\]

Assuming the $k$-th subset $\mathcal{P}_k$ contains $N_k$ points, and we want to upsample it into a subset $\mathcal{P}_k'$ containing $N_k'$ points, where $N_k'>N_k$.

During the upsampling process, use graph convolutional networks to process each subset. Assuming a graph convolutional network has $L$ layers of convolution, and the weight matrix of each layer is $\mathbf {W}_l$, then the output of layer $l$ is:

\[
\mathbf{H}^{(l)} = \sigma(\mathbf{A} \mathbf{H}^{(l-1)} \mathbf{W}^{(l)})
\]

The pseudocode for the above method is showing in Algorithm ~\ref{alg:1}.
\noindent
\begin{figure}[t]
\centering
\scalebox{0.7}{
\begin{minipage}[t]{0.85\textwidth}
    \vspace{0pt}
    \small
    \begin{algorithm}[H]
    \caption{Point Cloud Upsampling with Patch Segmentation}\label{alg:1}
    \KwIn{Data matrix $\mathbf{data}$ of shape $(N, P, 3)$, number of clusters $K$\\ \textcolor{gray}{\#$N \gets \text{number of objects} \gets \mathbf{data}.\text{shape}[0]$}\\ \textcolor{gray}{\#$P \gets \text{number of points} \gets \mathbf{data}.\text{shape}[1]$}}
    
    $\text{segmented\_data} \gets \text{split}(\mathbf{data}, K, \text{axis}=0)$

    \For{each segment in $\text{segmented\_data}$}{$\text{segment} \gets \text{reshape}(\text{segment}, (-1, \frac{P}{K}, 3))$
    }
    \end{algorithm}
\end{minipage}
}
\end{figure}

\section{Experiments}
\subsection{Settings}
\textbf{Datasets:} In general speaking, the experiment in this article applied the following two 3D benchmark datasets:

\begin{itemize}
 \item \textbf{PU1K:} 
This is a point cloud upsampling dataset introduced in PU-GCN \cite{qian2021pugcn}. Overall, PU1K consists of 1147 3D models, including 120 3D models from the PU-GAN dataset, as well as 900 different models collected from ShapeNetCore \cite{chang2015shapenet}. The test set includes 27 models from PU-GAN and over 100 models from ShapeNetCore. Covering 50 object categories.

\end{itemize}

\begin{itemize}
 \item \textbf{ABC Dataset:} 
The ABC Dataset \cite{koch2019abc} is a dataset containing one million computer-aided design (CAD) models, aimed at promoting research on geometric deep learning methods and applications. It is currently one of the largest publicly available datasets for geometric deep learning research. 

\end{itemize}

\begin{figure}[tb]
  \centering
  \includegraphics[width=\textwidth]{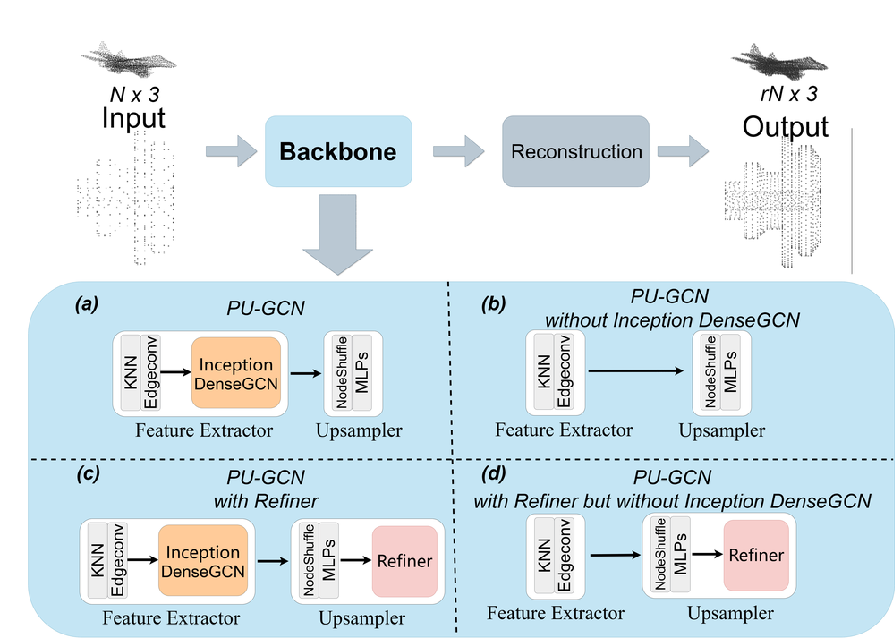}
  \caption{The PU-GCN network structure used in the experiment is shown in the figure, where (a) is the original PU-GCN structure. (b), (c) and (d) are all architectures used in the ablation experiment to explore the modules that have the greatest impact on upsampling.
  }
  \label{fig:3}
\end{figure}

\textbf{Training details:} Straightforwardly, the PU-GCN used in this article, where the network structure is shown in the architecture of (a) in Figure ~\ref{fig:3}, is implemented using PyTorch. Due to TensorFlow and CUDA versions, this article is unable to re-create the original author's Python environment in PU-GCN within the TensorFlow deep learning framework. And the experiments were conducted on a single NVIDIA GeForce RTX 3060 LHR GPU running on a Ubuntu 22.04 operating system. In terms of training hyperparameters, this article follows the settings of the original PU-GCN in terms of network structure. Notwithstanding, a batch size of $32$ with $50$ training epochs was used, with an initial learning rate of $5\times 10^{-4}$, a decay rate of $0.05$, and so on.

\subsection{Evaluation Metrics}
During the testing process, this article adopted the common practices used in previous work on point cloud upsampling \cite{qian2021pugcn,qiu2022putransformer,zhao2022selfsupervised,he2023gradpu,kim2024puedgeformer,mao2022pu}. That is to say, this article will cut the input point cloud into multiple subset Patch and AS of all N points. 

In the 4 times upsampling experiment of this article, three types of low resolution to high-resolution sampling were performed. Each test sample has a low resolution point cloud consisting of 256, 512 and 1048 points, and a high-resolution point cloud consisting of 1024,2048 and 8196 points. This article quantitatively evaluates the upsampling performance of PU-GCN under two methods based on three widely used indicators: \emph{(i)} Chamfer distance (CD) \cite{hajdu2012approximations}, \emph{(ii)} Hausdorff distance (HD) \cite{berger2013benchmark}, and \emph{(iii)} model parameter quantity.

\subsection{Results}
\noindent \textbf{PU1K dataset:} Table ~\ref{Tab:1} shows the quantitative results of using patch method and uniform segmentation method with PU-GCN network training on the PU1K dataset. By upsampling at resolutions of 256-1024 and 2048-8192, it can be seen that the Patch based method outperforms the AS method proposed in this paper in all two metrics of distance measurement.

\begin{table}[]
\centering
\caption{The data input based on the Patch method and AS method is obtained using PU-GCN in PU1K.}
\label{Tab:1}
\scalebox{0.73}{
\begin{tabular}{ccccc}
\hline
\hline
\multirow{2}{*}{Datasets}                                                       & \multirow{2}{*}{Test Scale}     & \multirow{2}{*}{Method} & \multicolumn{2}{c}{Results $(\times 10^{-3})$}         \\ \cline{4-5} 
                                                                                &                                 &                         & \multicolumn{1}{c|}{CD $\downarrow$} & HD $\downarrow$ \\ \hline
\multirow{6}{*}{\begin{tabular}[c]{@{}c@{}}PU1K\\ $256\to 1024$\end{tabular}}   & \multirow{2}{*}{$256\to 1024$}  & Patch                   & \textbf{35.459}                      & \textbf{54.884} \\
                                                                                &                                 & AS                      & 35.529                               & 54.923          \\ \cline{2-5} 
                                                                                & \multirow{2}{*}{$512\to 2048$}  & Patch                   & 25.689                               & \textbf{41.990} \\
                                                                                &                                 & AS                      & \textbf{25.616}                      & 42.344          \\ \cline{2-5} 
                                                                                & \multirow{2}{*}{$2048\to 8192$} & Patch                   & \textbf{14.751}                      & \textbf{23.915} \\
                                                                                &                                 & AS                      & 14.961                               & 27.822          \\ \hline
\hline
\end{tabular}
}
\end{table}

\noindent \textbf{ABC10K dataset:} we also conducted point cloud upsampling experiments using the ABC10K dataset. Perform upsampling at a resolution of 512-2048. As shown in Table ~\ref{Tab:2}, in the 4$\times$ upsampling experiment, the two evaluation metrics of the Patch based method outperformed the AS method proposed in this paper in all two metrics of distance measurement.

\begin{table}[]
\centering
\caption{The data input based on the Patch method and AS method is obtained using PU-GCN in ABC10K.}
\label{Tab:2}
\scalebox{0.73}{
\begin{tabular}{ccccc}
\hline
\hline
\multirow{2}{*}{Datasets}                                                       & \multirow{2}{*}{Test Scale}     & \multirow{2}{*}{Method} & \multicolumn{2}{c}{Results $(\times 10^{-3})$}         \\ \cline{4-5} 
                                                                                &                                 &                         & \multicolumn{1}{c|}{CD $\downarrow$} & HD $\downarrow$ \\ \hline
\multirow{2}{*}{\begin{tabular}[c]{@{}c@{}}ABC10K\\ $512\to 2048$\end{tabular}} & \multirow{2}{*}{$512\to 2048$}  & Patch                   & \textbf{42.861}                      & \textbf{84.128} \\
                                                                                &                                 & AS                      & 47.991                               & 95.154          \\ \hline
\hline
\end{tabular}
}
\end{table}

Although article proposed name Average Segment (AS) method has many advantages in theory, experimental results show that the Patch based upsampling method performs better on Chamfer distance and Hausdorff distance. One possible reason for this may be that when upsampling on a patch, each patch contains fewer point clouds, and the graph convolutional network can extract local features more finely. Ergo, local features are easier to capture and reconstruct, resulting in excellent performance on CD and HD.

\subsection{Ablations}
\noindent \textbf{Test sets cross validation:} This article will conduct cross validation on the PU-GCN trained on the PU1K and ABC10K datasets under the validation sets of both datasets. To understand the impact of the model used for training data on upsampling results. The results are shown in Table ~\ref{Tab:3}. When the Patch method and AS method are trained on the PU1K dataset and tested on the ABC10K dataset, the CD and HD metrics are significantly superior to those trained and tested on the ABC10K dataset. However, when trained on the ABC10K dataset and tested on the PU1K dataset, there was no significant difference between the CD and HD indicators.

\begin{table}[]
\centering
\caption{The results of swapping test sets based on Patch method and AS method on two datasets.}
\label{Tab:3}
\scalebox{0.68}{
\begin{tabular}{ccccc}
\hline
\hline
\multirow{2}{*}{\begin{tabular}[c]{@{}c@{}}Training\\ Sets\end{tabular}}                                                  & \multirow{2}{*}{Method} & \multirow{2}{*}{Test Sets}                                                      & \multicolumn{2}{c}{Results $(\times 10^{-3})$} \\ \cline{4-5} 
                                                                                &                         &                                                                                 & CD $\downarrow$        & HD $\downarrow$       \\ \hline
\multirow{2}{*}{\begin{tabular}[c]{@{}c@{}}PU1K\\ $256\to 1024$\end{tabular}}   & Patch                   & \multirow{2}{*}{\begin{tabular}[c]{@{}c@{}}ABC10K\\ $512\to 2048$\end{tabular}} & \textbf{22.300}        & \textbf{43.386}       \\
                                                                                & AS                      &                                                                                 & 41.286                 & 82.695                \\ \hline
\multirow{6}{*}{\begin{tabular}[c]{@{}c@{}}ABC10K\\ $512\to 2048$\end{tabular}} & Patch                   & \multirow{2}{*}{\begin{tabular}[c]{@{}c@{}}PU1K\\ $256\to 1024$\end{tabular}}   & \textbf{35.260}        & \textbf{53.203}       \\
                                                                                & AS                      &                                                                                 & 38.494                 & 57.212                \\ \cline{2-5} 
                                                                                & Patch                   & \multirow{2}{*}{\begin{tabular}[c]{@{}c@{}}PU1K\\ $512\to 2048$\end{tabular}}   & \textbf{25.200}        & \textbf{39.345}       \\
                                                                                & AS                      &                                                                                 & 29.225                 & 43.932                \\ \cline{2-5} 
                                                                                & Patch                   & \multirow{2}{*}{\begin{tabular}[c]{@{}c@{}}PU1K\\ $2048\to 8192$\end{tabular}}  & \textbf{13.121}        & \textbf{21.707}       \\
                                                                                & AS                      &                                                                                 & 18.689                 & 27.688                \\ \hline
\hline
\end{tabular}
}
\end{table}

\noindent \textbf{Architecture Impact:} This article breaks down the important modules of PU-GCN, aiming to explore which part has the greatest impact on point cloud upsampling. To this end, the original network backbone has been increased or decreased in three ways: removing DenseGCN \emph{(b)}, adding Refiner \emph{(c)}, and removing DenseGCN while adding Refiner \emph{(d)}, all of them are showed in Figure ~\ref{fig:3}. The experiment was trained on the PU1K dataset and tested on the test sets of the PU1K dataset and ABC10K dataset. The results are shown in Table ~\ref{Tab:4}. The results showed that upsampling on the test set with three resolutions of 256-1024, 512-2048 and 2048-8192 in PU1K performed better than the original PU-GCN with DenseGCN, regardless of the presence or absence of Refiner, under the condition of removing DenseGCN. Under the upsampling scale of 2048-8192, the uniform segmentation method that removes DenseGCN is better than the Patch method.

\begin{table}[]
\centering
\caption{Perform ablation experiments on different architectures of PU-GCN using two data input methods.}
\label{Tab:4}
\scalebox{0.68}{
\begin{tabular}{cccccc}
\hline
\hline
\multirow{2}{*}{\begin{tabular}[c]{@{}c@{}}Network\\ Backbone\end{tabular}}                                & \multirow{2}{*}{Method} & \multirow{2}{*}{Test Sets}                                                     & \multicolumn{2}{c}{Results $(\times 10^{-3})$}      & \multirow{2}{*}{Model size (MB)} \\ \cline{4-5}
                                                                                                           &                         &                                                                                & CD $\downarrow$          & HD $\downarrow$          &                                  \\ \hline
\multirow{4}{*}{PU-GCN}                                                                                    & Patch                   & \multirow{2}{*}{\begin{tabular}[c]{@{}c@{}}PU1K\\ $256\to 1024$\end{tabular}}  & \textbf{35.459}          & \textbf{54.884}          & \multirow{4}{*}{\textbf{66.246}} \\
                                                                                                           & AS                      &                                                                                & 35.529                   & 54.923                   &                                  \\ \cline{2-5}
                                                                                                           & Patch                   & \multirow{2}{*}{\begin{tabular}[c]{@{}c@{}}PU1K\\ $2048\to 8192$\end{tabular}} & \textbf{14.751}          & \textbf{23.915}          &                                  \\
                                                                                                           & AS                      &                                                                                & 14.961                   & 27.822                   &                                  \\ \hline
\multirow{4}{*}{\begin{tabular}[c]{@{}c@{}}w/o \\ Inception DenseGCN\end{tabular}}                     & Patch                   & \multirow{2}{*}{\begin{tabular}[c]{@{}c@{}}PU1K\\ $256\to 1024$\end{tabular}}  & \textbf{35.061}          & 54.340                   & \multirow{4}{*}{99.152}          \\
                                                                                                           & AS                      &                                                                                & 35.229                   & \textit{\textbf{52.819}} &                                  \\ \cline{2-5}
                                                                                                           & Patch                   & \multirow{2}{*}{\begin{tabular}[c]{@{}c@{}}PU1K\\ $2048\to 8192$\end{tabular}} & 13.736                   & \textbf{22.573}          &                                  \\
                                                                                                           & AS                      &                                                                                & \textbf{13.504}          & 23.130                   &                                  \\ \hline
\multirow{4}{*}{w/ Refiner}                                                                              & Patch                   & \multirow{2}{*}{\begin{tabular}[c]{@{}c@{}}PU1K\\ $256\to 1024$\end{tabular}}  & \textbf{35.075}          & 54.227                   & \multirow{4}{*}{87.773}          \\
                                                                                                           & AS                      &                                                                                & 35.860                   & \textbf{53.652}          &                                  \\ \cline{2-5}
                                                                                                           & Patch                   & \multirow{2}{*}{\begin{tabular}[c]{@{}c@{}}PU1K\\ $2048\to 8192$\end{tabular}} & \textbf{14.361}          & \textbf{23.275}          &                                  \\
                                                                                                           & AS                      &                                                                                & 14.817                   & 26.860                   &                                  \\ \hline
\multirow{4}{*}{\begin{tabular}[c]{@{}c@{}}w/ Refiner \\ w/o Inception DenseGCN\end{tabular}} & Patch                   & \multirow{2}{*}{\begin{tabular}[c]{@{}c@{}}PU1K\\ $256\to 1024$\end{tabular}}  & \textit{\textbf{34.309}} & 54.070                   & \multirow{4}{*}{130.945}         \\
                                                                                                           & AS                      &                                                                                & 35.158                   & \textbf{53.434}          &                                  \\ \cline{2-5}
                                                                                                           & Patch                   & \multirow{2}{*}{\begin{tabular}[c]{@{}c@{}}PU1K\\ $2048\to 8192$\end{tabular}} & 13.721                   & \textit{\textbf{22.245}} &                                  \\
                                                                                                           & AS                      &                                                                                & \textit{\textbf{13.441}} & 22.899                   &                                  \\ \hline
\hline
\end{tabular}
}
\end{table}

\subsection{Visualizations}
\begin{figure}[tb]
  \centering
  \includegraphics[height=3.6cm]{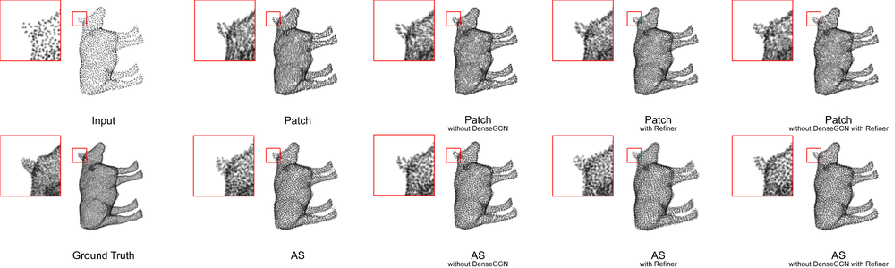}
  \caption{Visualization of PU-GCN with Different Architectures on the PU1K Dataset.
  }
  \label{fig:4}
\end{figure}

\begin{figure}[tb]
  \centering
  \includegraphics[height=3.6cm]{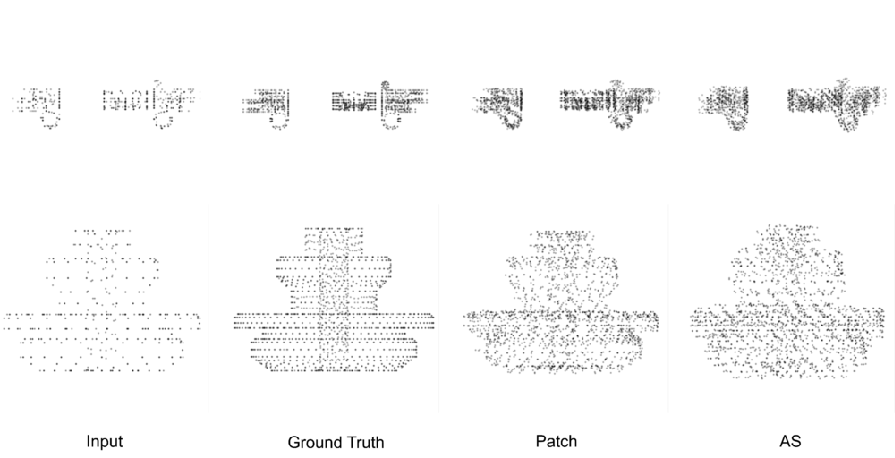}
  \caption{Visualization example on the ABC dataset.
  }
  \label{fig:5}
\end{figure}

The qualitative results of upsampling using different point cloud datasets are shown in Figure ~\ref{fig:4} and Figure ~\ref{fig:5}. It can be seen that the input based on Patch performs better in distinguishing edges on the boundaries compared to the AS method. This is particularly evident on the CAD based ABC dataset, as shown in Figure ~\ref{fig:5}. However, in Figure ~\ref{fig:4}, it can be seen that the AS method performs better locally in the model. Relatively speaking, it is denser and the connections between points are tighter. In addition, even though DenseGCN performs better on CD and HD on the PU1K dataset, this results in the point cloud lacking local features in visualization.

\section{Discussion}
Now, we will focus on why the patch upsampling method performs better in PU-GCN experiments. Although the AS random sampling method proposed in this article has many advantages, the actual experimental results are not as good as the input method based on Patch in terms of Chamfer distance (CD) and Hausdorff distance (HD). This phenomenon can be explained by the following points:

\textbf{The effectiveness of boundary processing strategies:} Although theoretically AS sampling followed by segmentation can avoid boundary problems, in practical applications, the overlapping and smoothing of boundaries in patch segmentation may be more effective than those after AS sampling, thereby improving the overall quality of upsampling.

Set two adjacent patches as $\mathcal{P}_k$ and $\mathcal{P}_m$. The overlapping area between them is $\mathcal{O}_{k,m}$, whose point cloud is represented as $\mathbf{X}_{k,m} \in \mathbb{R}^{N_{k,m} \times d}$, where $N_{k,m}$ is the number of overlapping regions.

Smoothing in overlapping areas can reduce errors caused by boundary effects:

\[
\mathbf{X}_{k,m}' = \alpha \mathbf{X}_{k,m}^{(k)} + (1 - \alpha) \mathbf{X}_{k,m}^{(m)}
\]

where:
\begin{itemize}
    \item $\mathbf{X}_{k,m}^{(k)}$ and $\mathbf{X}_{k,m}^{(m)}$ respectively represent overlapping area point clouds in Patch $\mathcal{P}_k$ and $\mathcal{P}_m$.
    
    \item $\alpha \in [0,1]$ is the smoothing coefficient
\end{itemize}

The smoothing of this overlapping area can effectively reduce boundary effects and improve the overall quality of sampling.

\textbf{Insufficient utilization of global information:} Although AS methods retain global information, some global information may be lost during processing and segmentation, resulting in a decrease in the quality of the upsampling point cloud. On the contrary, although the patch method processes local information, it can still reconstruct global information well through the synthesis of multiple patches.

The method for AS as a whole is $\mathcal{P}_{\text{sampled}} = \{\mathbf{p}_i \mid i \in I_{\text{sampled}}\}$, divide it into $K$ subsets:

\[
\mathcal{P}_k = \{\mathbf{p}_i \mid i \in I_k, I_k \subset I_{\text{sampled}}, |I_k| = \frac{M}{K}\}
\]

Due to the segmented subset $\mathcal{P}_k$ lack of local consistency may lead to a decrease in the quality of feature extraction and reconstruction.

\section{Future Works}
PU-GCN model used in this article as an example, the existing Patch based upsampling method is better than AS method we proposed. This may be due to the difference in dataset size between the two methods. In future work, we will delve deeper into the differences in interpretability between Patch and AS methods through representation learning methods. At the same time, due to the similarity between the use of point clouds and the completion of point clouds. We will refer to the method in article ~\cite{xia2021asfm} and use encoders for the data input of Patch and AS respectively, in order to merge global and edge features and achieve better results.

\section{Conclusion}
To ensure the model integrity of point cloud upsampling, this article mainly studies the input of point cloud data. Specifically, this article proposes a new data input method, also known as a new training model method, called Average Segment. And experiments were conducted on a stable and lightweight PU-GCN model. The traditional patch method still performs slightly better in both CD and HD indicators, but through relevant ablation research and visualization, we can see that the AS method can maintain local features within the model.

%
%
\bibliographystyle{unsrt}
\bibliography{main}
\end{document}